\documentclass[runningheads]{llncs}

\usepackage{graphicx}
\usepackage{textcomp}
\usepackage{wrapfig}
\usepackage{subcaption}
\usepackage{verbatim}
\usepackage{enumitem}
\begin{document}
\setlist[itemize]{noitemsep, topsep=1pt}
\title{Resolving the Scope of Speculation and Negation using Transformer-Based Architectures}
\titlerunning{Resolving Scope of Spec. and Neg. using Transformer-Based Arch.}
\author{Benita Kathleen Britto\inst{1} \and
Aditya Khandelwal\inst{2}}
\authorrunning{B.K. Britto, A. Khandelwal}
\institute{Veermata Jijabai Technological Institute, Mumbai \\ \email{bcbritto\_b16@it.vjti.ac.in}
\and
College of Engineering Pune \\
\email{khandelwalar16.comp@coep.ac.in}}
\maketitle              % typeset the header of the contribution
\begin{abstract}
Speculation is a naturally occurring phenomena in textual data, forming an integral component of many systems, especially in the biomedical information retrieval domain. Previous work addressing cue detection and scope resolution (the two subtasks of speculation detection) have ranged from rule-based systems to deep learning-based approaches. In this paper, we apply three popular transformer-based architectures, BERT, XLNet and RoBERTa to this task, on two publicly available datasets, BioScope Corpus and SFU Review Corpus, reporting substantial improvements over previously reported results (by at least 0.29 F1 points on cue detection and 4.27 F1 points on scope resolution). We also experiment with joint training of the model on multiple datasets, which outperforms the single dataset training approach by a good margin. We observe that XLNet consistently outperforms BERT and RoBERTa, contrary to results on other benchmark datasets. To confirm this observation, we apply XLNet and RoBERTa to negation detection and scope resolution, reporting state-of-the-art results on negation scope resolution for the BioScope Corpus (increase of 3.16 F1 points on the BioScope Full Papers, 0.06 F1 points on the BioScope Abstracts) and the SFU Review Corpus (increase of 0.3 F1 points).

\keywords{Speculation Scope Resolution \and Negation Scope Resolution \and Transformers}
\end{abstract}
\section{Introduction}
\noindent
\par
\begin{figure}[!htb]
    \centering
    \includegraphics[width=0.99\textwidth]{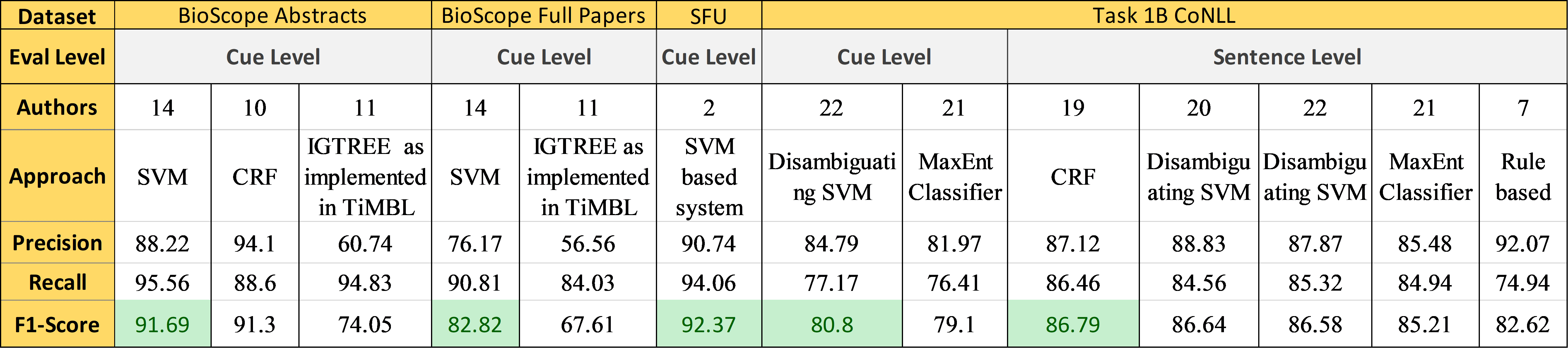}
    \caption{Literature Review: Speculation Cue Detection}
    \label{fig:litreviewcue}

    \centering
    \includegraphics[width=0.99\textwidth]{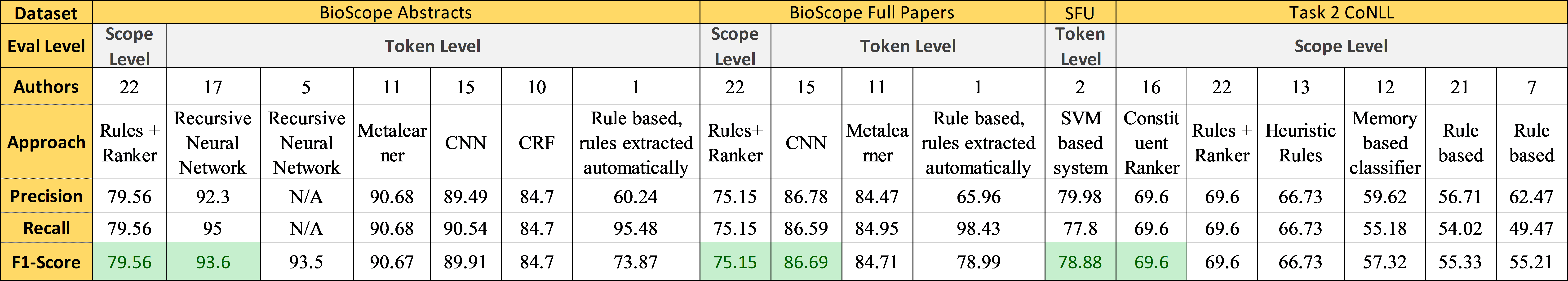}
    \caption{Literature Review: Speculation Scope Resolution}
    \label{fig:litreviewscope}
\end{figure}

 The task of speculation detection and scope resolution is critical in distinguishing factual information from speculative information. This has multiple use-cases, like systems that determine the veracity of information, and those that involve requirement analysis. This task is particularly important to the biomedical domain, where patient reports and medical articles often use this feature of natural language. This task is commonly broken down into two subtasks: the first subtask, speculation cue detection, is to identify the uncertainty cue in a sentence, while the second subtask: scope resolution, is to identify the scope of that cue. For instance, consider the example: \\ \centerline{\begin{em}It might rain tomorrow.\end{em}}
\indent The speculation cue in the sentence above is ‘might’ and the scope of the cue \begin{em}‘might’\end{em} is \begin{em}‘rain tomorrow’\end{em}. Thus, the speculation cue is the word that expresses the speculation, while the words affected by the speculation are in the scope of that cue.
\\
\indent This task was the CoNLL-2010 Shared Task (\cite{farkas-etal-2010-conll}), which had 3 different subtasks. Task 1B was speculation cue detection on the BioScope Corpus, Task 1W was weasel identification from Wikipedia articles, and Task 2 was speculation scope resolution from the BioScope Corpus. For each task, the participants were provided the train and test set, which is henceforth referred to as Task 1B CoNLL and Task 2 CoNLL throughout this paper.
\\
\indent For our experimentation, we use the sub corpora of the BioScope Corpus (\cite{szarvas-etal-2008-bioscope}), namely the BioScope Abstracts sub corpora, which is referred to as BA, and the BioScope Full Papers sub corpora, which is referred to as BF. We also use the SFU Review Corpus (\cite{konstantinova-etal-2012-review}), which is referred to as SFU.
\\
\indent This subtask of natural language processing, along with another similar subtask, negation detection and scope resolution, have been the subject of a body of work over the years. The approaches used to solve them have evolved from simple rule-based systems (\cite{kilicoglu-bergler-2010-high}) based on linguistic information extracted from the sentences, to modern deep-learning based methods. The Machine Learning techniques used varied from Maximum Entropy Classifiers (\cite{velldal-etal-2010-resolving}) to Support Vector Machines (\cite{diaz-noa-taboada},\cite{ozgur-radev-2009-detecting},\cite{velldal},\cite{velldal-etal-2012-speculation}), while the deep learning approaches included Recursive Neural Networks (\cite{FEI202022},\cite{ren-yafeng}), Convolutional Neural Networks (\cite{qian-etal-2016-speculation}) and most recently transfer learning-based architectures like Bidirectional Encoder Representation from Transformers (BERT) (\cite{2019arXiv191104211K}). Figures \ref{fig:litreviewcue} and \ref{fig:litreviewscope} contain a summary of the papers addressing speculation detection and scope resolution  (\cite{apostolova-etal-2011-automatic}, \cite{diaz-noa-taboada}, \cite{FEI202022}, \cite{kilicoglu-bergler-2010-high}, \cite{moncecchi-etal-2012-improving}, \cite{morante-daelemans-2009-learning}, \cite{morante-etal-2010-memory}, \cite{ovrelid-etal-2010-syntactic}, \cite{ozgur-radev-2009-detecting}, \cite{qian-etal-2016-speculation}, \cite{read-jonathon-velldal}, \cite{ren-yafeng}, \cite{tang-etal-2010-cascade}, \cite{velldal},  \cite{velldal-etal-2010-resolving}, \cite{velldal-etal-2012-speculation}). \par

\par Inspired by the most recent approach of applying BERT to negation detection and scope resolution (\cite{2019arXiv191104211K}), we take this approach one step further by performing a comparative analysis of three popular transformer-based architectures: BERT (\cite{DBLP:journals/corr/abs-1810-04805}), XLNet (\cite{DBLP:journals/corr/abs-1906-08237}) and RoBERTa (\cite{DBLP:journals/corr/abs-1907-11692}), applied to speculation detection and scope resolution. We evaluate the performance of each model across all datasets via the single dataset training approach, and report all scores including inter-dataset scores (i.e. train on one dataset, evaluate on another) to test the generalizability of the models. This approach outperforms all existing systems on the task of speculation detection and scope resolution. Further, we jointly train on multiple datasets and obtain improvements over the single dataset training approach on most datasets.\par
Contrary to results observed on benchmark GLUE tasks, we observe XLNet consistently outperforming RoBERTa. To confirm this observation, we apply these models to the negation detection and scope resolution task, and observe a continuity in this trend, reporting state-of-the-art results on three of four datasets on the negation scope resolution task.\par
The rest of the paper is organized as follows: In Section 2, we provide a detailed description of our methodology and elaborate on the experimentation details. In Section 3, we present our results and analysis on the speculation detection and scope resolution task, using the single dataset and the multiple dataset training approach. In Section 4, we show the results of applying XLNet and RoBERTa on negation detection and scope resolution and propose a few reasons to explain why XLNet performs better than RoBERTa. Finally, the future scope and conclusion is mentioned in Section 5.

\section{Methodology and Experimental Setup}
\setlength{\parindent}{5ex}
\par We use the methodology by Khandelwal and Sawant (\cite{2019arXiv191104211K}), and modify it to support experimentation with multiple models.\\
\\
\textbf{For Speculation Cue Detection:}

Input Sentence: \begin{em}It might rain tomorrow.\end{em}

True Labels: \begin{em}Not-A-Cue, Cue, Not-A-Cue, Not-A-Cue.\end{em}

\noindent 
First, this sentence is preprocessed to get the target labels as per the following annotation schema:

    \begin{em}1 – Normal Cue \quad 2 – Multiword Cue \quad 3 – Not a cue \quad 4 – Pad token\end{em}

\noindent 
Thus, the preprocessed sequence is as follows:

Input Sentence: \quad \begin{em}[It, might, rain, tomorrow]\end{em}

True Labels: \quad \begin{em}[3,1,3,3]\end{em}

\noindent
Then,  we  preprocess  the  input  using the  tokenizer  for  the  model  being  used (BERT, XLNet or RoBERTa): splitting each word into one or more tokens, and converting each token to it’s corresponding tokenID, and padding it to the maximum input length of the model. Thus,

Input Sentence: \quad \begin{em}[wtt(It), wtt(might), wtt(rain),  wtt(tom), wtt(\#\# or), wtt(\#\# row), wtt(\textlangle pad \textrangle),wtt(\textlangle pad \textrangle)...]\end{em}

True Labels: \quad \begin{em}[3,1,3,3,3,3,4,4,4,4,...]\end{em}
\noindent
\\
The word ‘tomorrow' has been split into 3 tokens, ‘tom', ‘\#\#or' and ‘\#\#row'. The function to convert the word to tokenID is represented by wtt.

\noindent
\\
\textbf{For Speculation Scope Resolution:}

\noindent If a sentence has multiple cues, each cue's scope will be resolved individually.

\begin{wrapfigure}{r}{0.38\textwidth}
    \centering
    \includegraphics[width=0.38\textwidth]{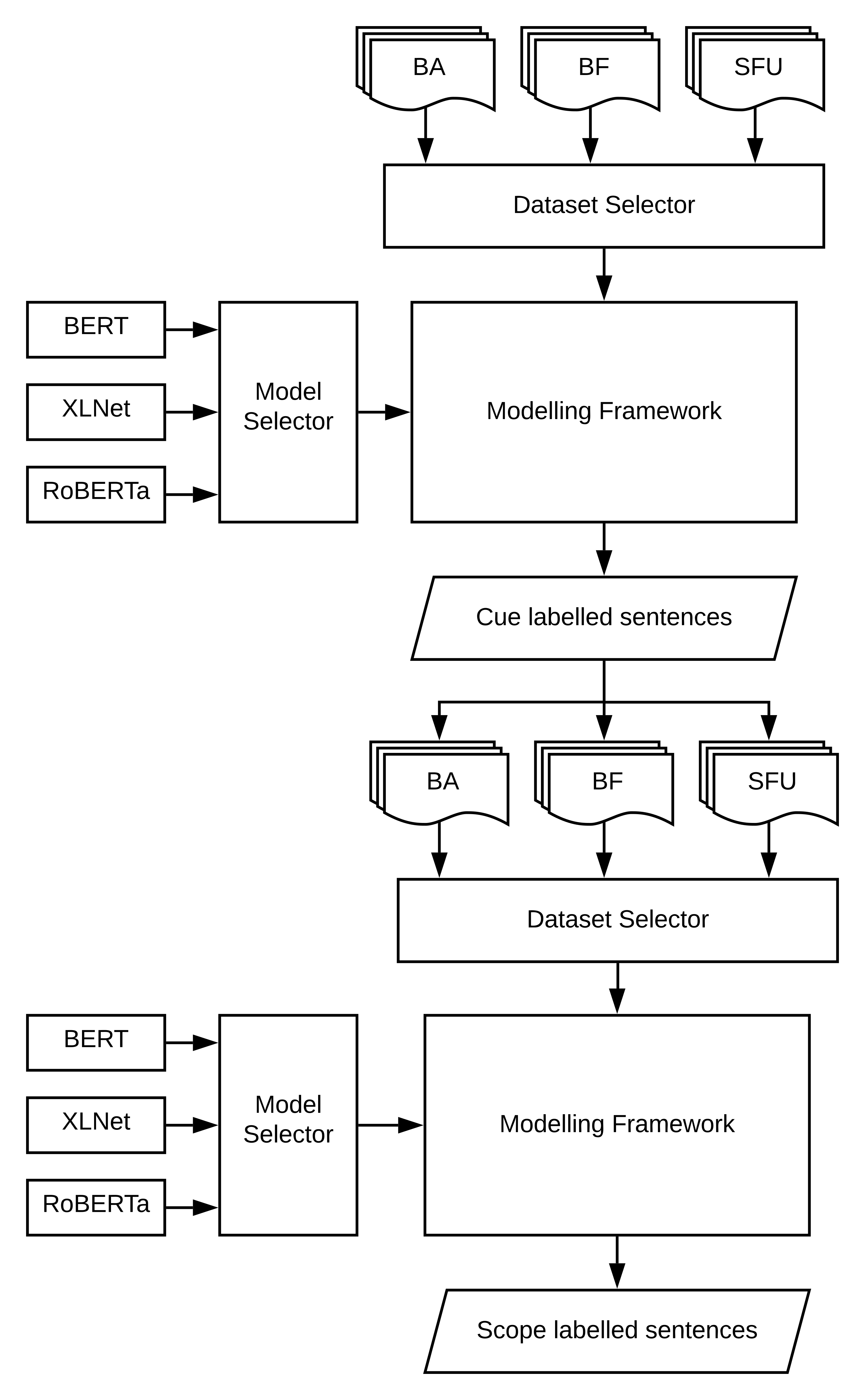}
    \caption{Our Approach}
\end{wrapfigure}

Input Sentence: \begin{em}It might rain tomorrow.\end{em}

True Labels: \begin{em}Out-Of-Scope, Out-Of-Scope, In-Scope, In-Scope.\end{em}

\noindent 
First, this sentence is preprocessed to get the target labels as per the following annotation schema: 
\noindent
\begin{em}0 – Out-Of-Scope \quad 1 – In-Scope\end{em}

\noindent 
Thus, the preprocessed sequence is as follows:

True Scope Labels: \quad \begin{em}[0,0,1,1]\end{em}

\noindent 
As for cue detection, we preprocess the input using the tokenizer for the model being used. Additionally, we need to indicate which cue's scope we want to find in the input sentence. We do this by inserting a special token representing the token type (according to the cue detection annotation schema) before the cue word whose scope is being resolved. Here, we want to find the scope of the cue ‘might'. Thus, 

Input Sentence: \quad \begin{em}[wtt(It), wtt(\textlangle token[1]\textrangle), wtt(might), wtt(rain), wtt(tom), wtt(\#\#or), wtt(\#\#row), wtt(\textlangle pad\textrangle),  wtt(\textlangle pad\textrangle)...]\end{em}

True Scope Labels: \quad \begin{em}[0,0,1,1,1,1,0,0,0,0,...]\end{em} 

\noindent Now, the preprocessed input for cue detection and similarly for scope detection is fed as input to our model as follows:\\
    \indent X = Model (Input)\\
    \indent Y = W*X + b
\par The W matrix is a matrix of size n\_hidden x num\_classes (n\_hidden is the size of the representation of a token within the model). These logits are fed to the loss function.
\noindent
We use the following variants of each model:
	\begin{itemize}
	    \item  BERT: bert-base-uncased\footnotemark  \footnotetext{s3.amazonaws.com/models.huggingface.co/bert/bert-base-uncased.tar.gz} (The model used by \cite{2019arXiv191104211K})
	    \item RoBERTa: roberta-base\footnotemark  \footnotetext{s3.amazonaws.com/models.huggingface.co/bert/roberta-base-pytorch\_model.bin} (RoBERTa-base does not have an uncased variant)
	    \item XLNet: xlnet-base-cased\footnotemark \footnotetext{s3.amazonaws.com/models.huggingface.co/bert/xlnet-base-cased-pytorch\_model.bin} (XLNet-base does not have an uncased variant)
	\end{itemize}

\noindent
The output of the model is a vector of probabilities per token. The loss is calculated for each token, by using the output vector and the true label for that token. We use class weights for the loss function, setting the weight for label 4 to 0 and all other labels to 1 (for cue detection only) to avoid training on padding token’s output.

\noindent \par We calculate the scores (Precision, Recall, F1) for the model per word of the input sentence, not per token that was fed to the model, as the tokens could be different for different models leading to inaccurate scores. For the above example, we calculate the output label for the word ‘tomorrow', not for each token it was split into (‘tom', ‘\#\#or' and ‘\#\#row'). To find the label for each word from the tokens it was split into, we experiment with 2 methods:
\begin{enumerate}
    \item Average: We average the output vectors (softmax probabilities) for each token that the word was split into by the model's tokenizer. In the example above, we average the output of ‘tom', ‘\#\#or' and ‘\#\#row' to get the output for ‘tomorrow'. Then, we take an argmax over the resultant vector. This is then compared with the true label for the original word.
    
    \item First Token: Here, we only consider the first token's probability vector (among all tokens the word was split into) as the output for that word, and get the label by an argmax over this vector. In the example above, we would consider the output vector corresponding to the token ‘tom' as the output for the word ‘tomorrow'. 
\end{enumerate}
\par For cue detection, the results are reported for the Average method only, while we report the scores for both Average and First Token for Scope Resolution.
\noindent
\par For fair comparison, we use the same hyperparameters for the entire architecture for all 3 models. Only the tokenizer and the model are changed for each model. All other hyperparameters are kept same. We finetune the models for 60 epochs, using early stopping with a patience of 6 on the F1 score (word level) on the validation dataset. We use an initial learning rate of 3e-5, with a batch size of 8. We use the Categorical Cross Entropy loss function. \par
We use the Huggingface’s Pytorch Transformer library (\cite{Wolf2019HuggingFacesTS}) for the models and train all our models on Google Colaboratory.

\section{Results: Speculation Cue Detection and Scope Resolution}
\begin{figure}[!htb]
\begin{subfigure}{0.42\textwidth}
    \includegraphics[width = 0.95\textwidth]{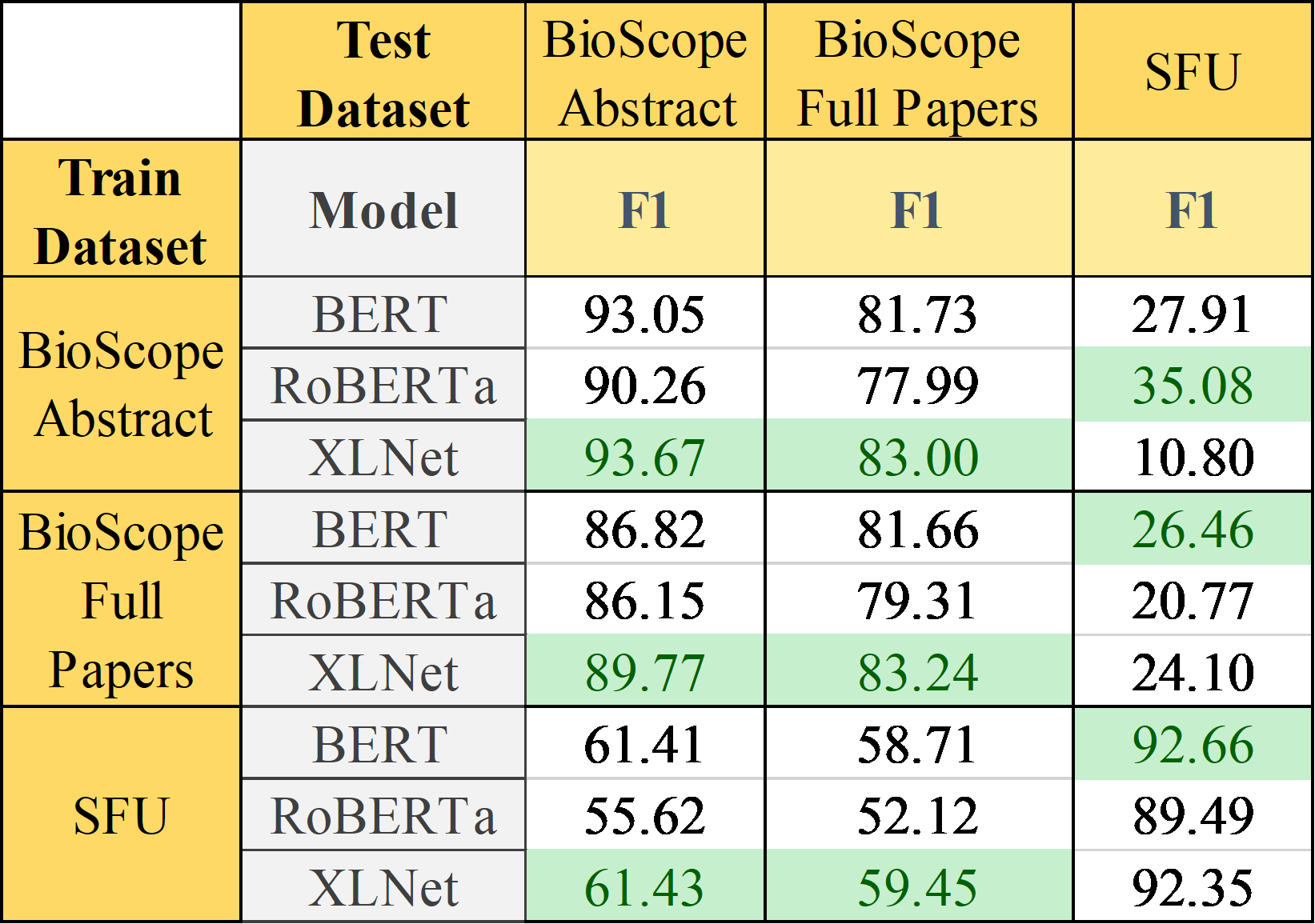}
    \caption{Speculation Cue Detection}
    \label{fig:speccue}
\end{subfigure}
\begin{subfigure}{0.58\textwidth}
    \includegraphics[width = 0.95\textwidth]{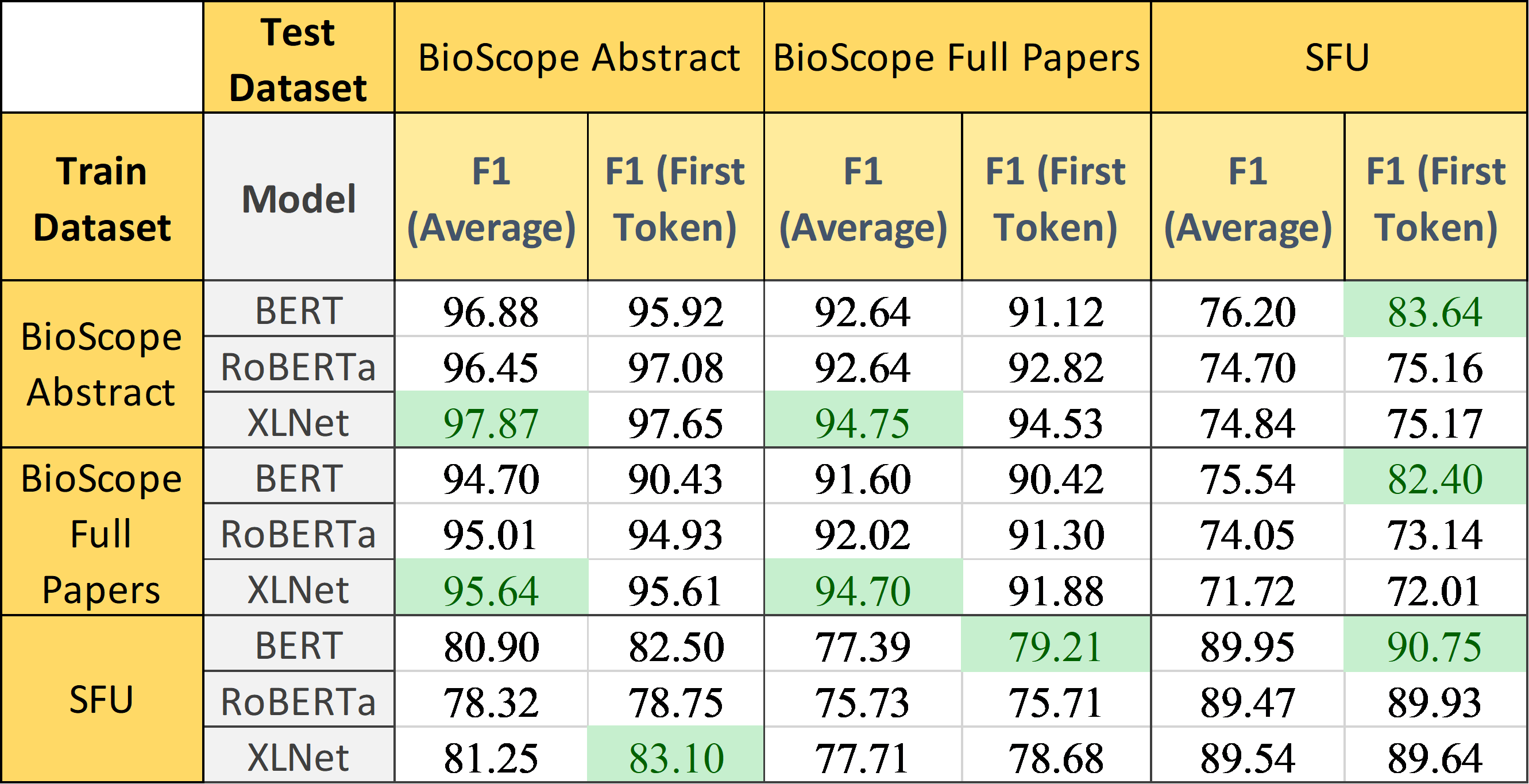}
    \caption{Speculation Scope Resolution}
    \label{fig:specscope}
\end{subfigure}

\caption{Results for Single Dataset Training}
\label{fig:spec}
\end{figure}
\begin{figure}[!htb]
\begin{subfigure}{0.43\textwidth}
    \includegraphics[width = 0.95\textwidth]{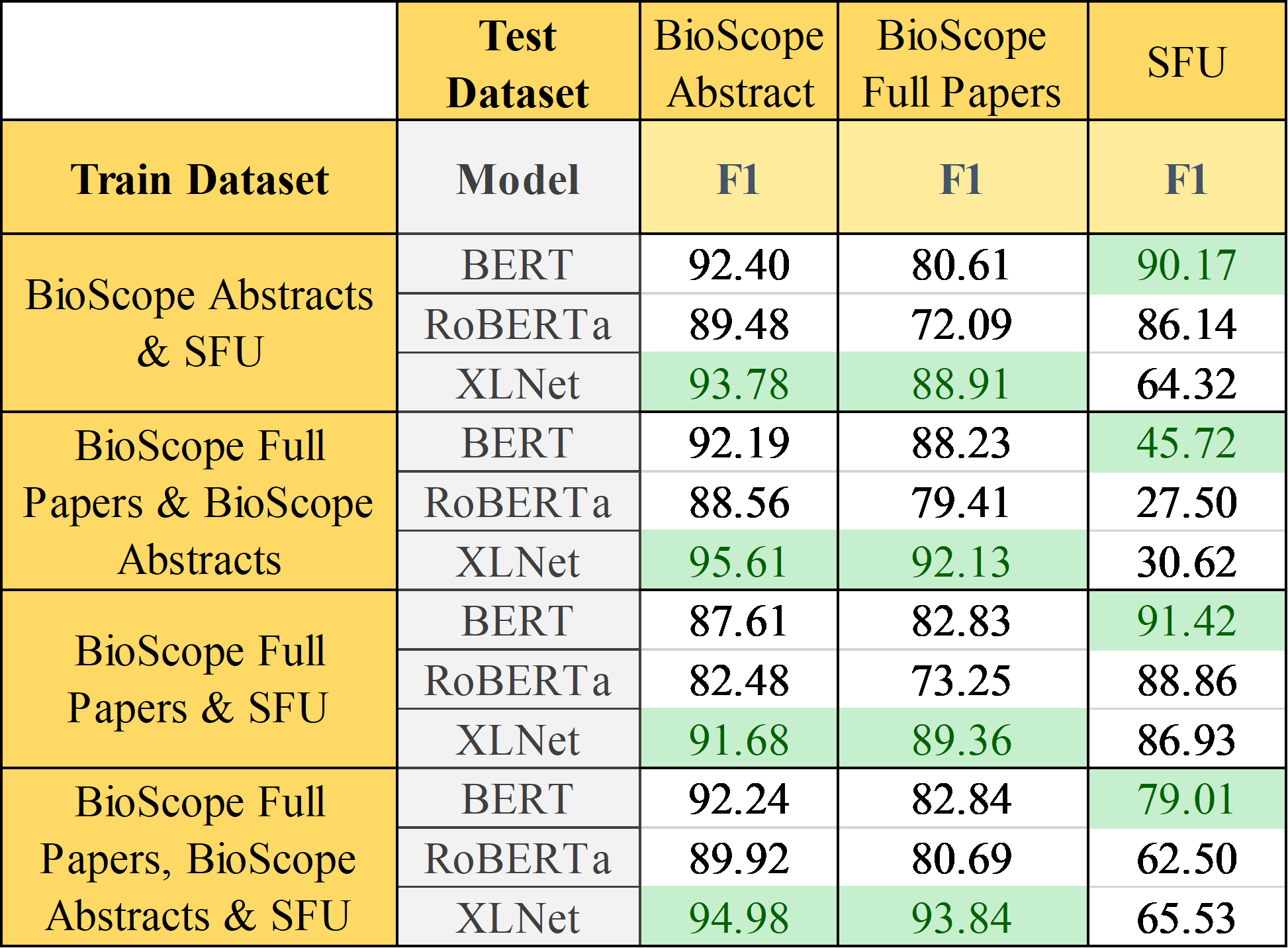}
    \caption{Speculation Cue Detection}
    \label{fig:multidataspeccue}
\end{subfigure}
\begin{subfigure}{0.57\textwidth}
    \includegraphics[width = 0.95\textwidth]{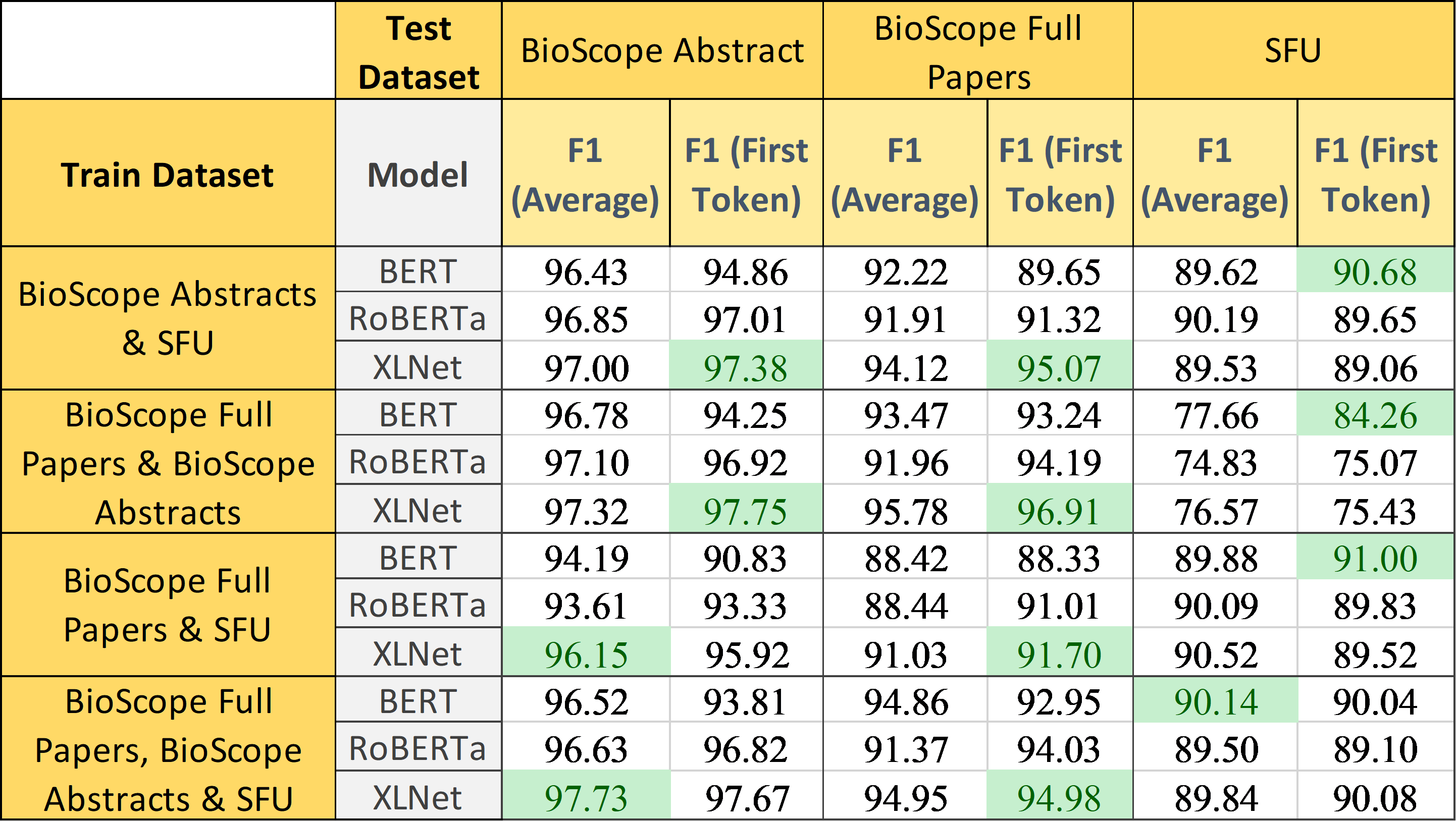}
    \caption{Speculation Scope Resolution}
    \label{fig:multidataspecscope}
\end{subfigure}

\caption{Results for Joint Training on Multiple Datasets}
\label{fig:multidata}
\end{figure}
\noindent
\par We use a default train-validation-test split of 70-15-15 for each dataset. For the speculation detection and scope resolution subtasks using single-dataset training, we report the results as an average of 5 runs of the model. For training the model on multiple datasets, we perform a 70-15-15 split of each training dataset, after which the train and validation part of the individual datasets are merged while the scores are reported for the test part of the individual datasets, which is not used for training or validation. We report the results as an average of 3 runs of the model. Figure \ref{fig:spec} contains results for speculation cue detection and scope resolution when trained on a single dataset.
All models perform the best when trained on the same dataset as they are evaluated on, except for BF, which gets the best results when trained on BA. This is because of the transfer learning capabilities of the models and the fact that BF is a smaller dataset than BA (BF: 2670 sentences, BA: 11871 sentences).
For speculation cue detection, there is lesser generalizability for models trained on BF or BA, while there is more generalizability for models trained on SFU. This could be because of the different nature of the biomedical domain.
\par Figure \ref{fig:multidata} contains the results for speculation detection and scope resolution for models trained jointly on multiple datasets. We observe that training on multiple datasets helps the performance of all models on each dataset, as the quantity of data available to train the model increases.
We also observe that XLNet consistently outperforms BERT and RoBERTa. To confirm this observation, we apply the 2 models to the related task of negation detection and scope resolution

\section{Negation Cue Detection and Scope Resolution}
\begin{figure}[!htb]
\begin{subfigure}{0.44\textwidth}
    \includegraphics[width = 0.95\textwidth]{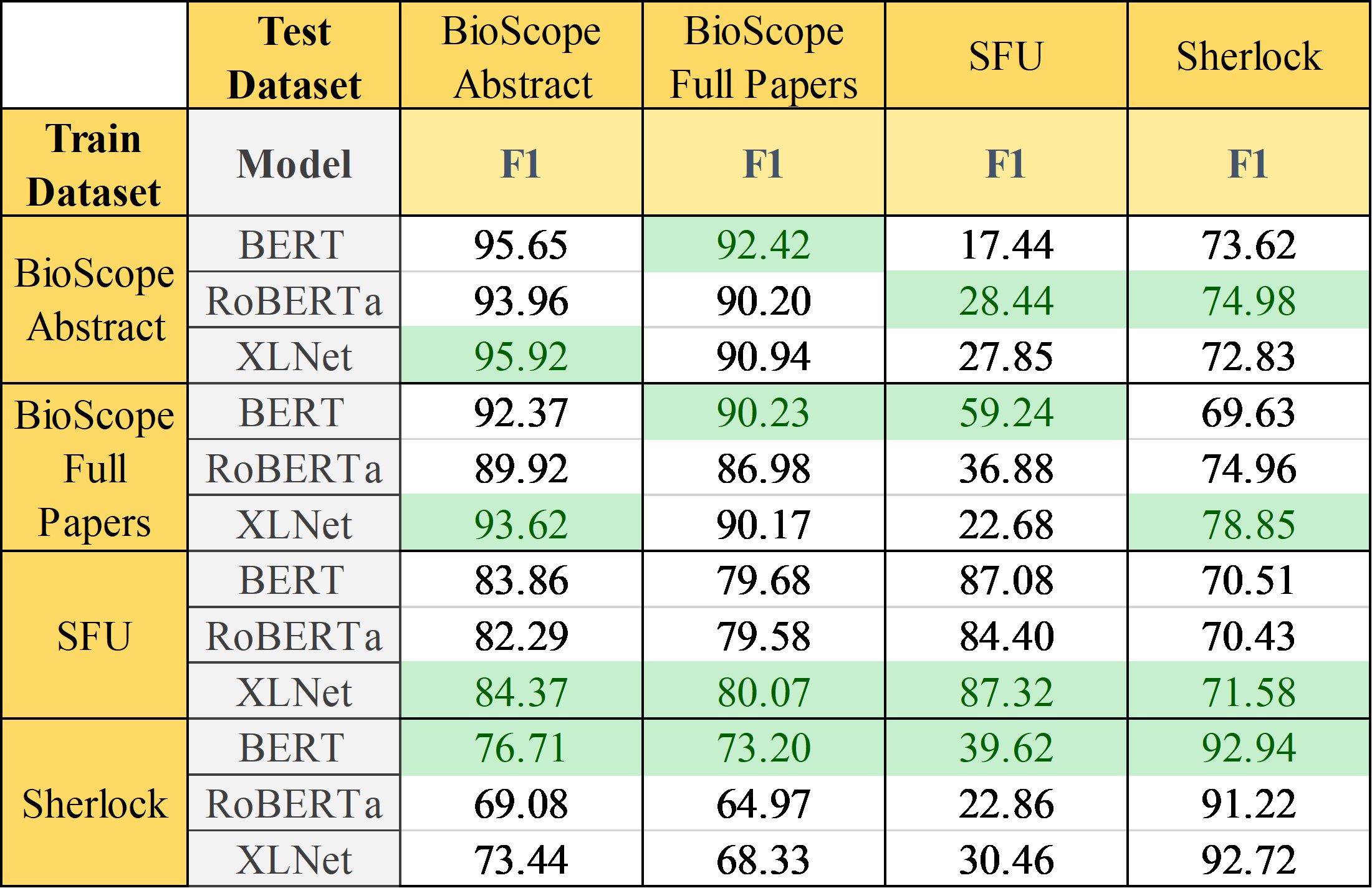}
    \caption{Negation Cue Detection}
    \label{fig:negcue}
\end{subfigure}
\begin{subfigure}{0.54\textwidth}
    \includegraphics[width = 0.95\textwidth]{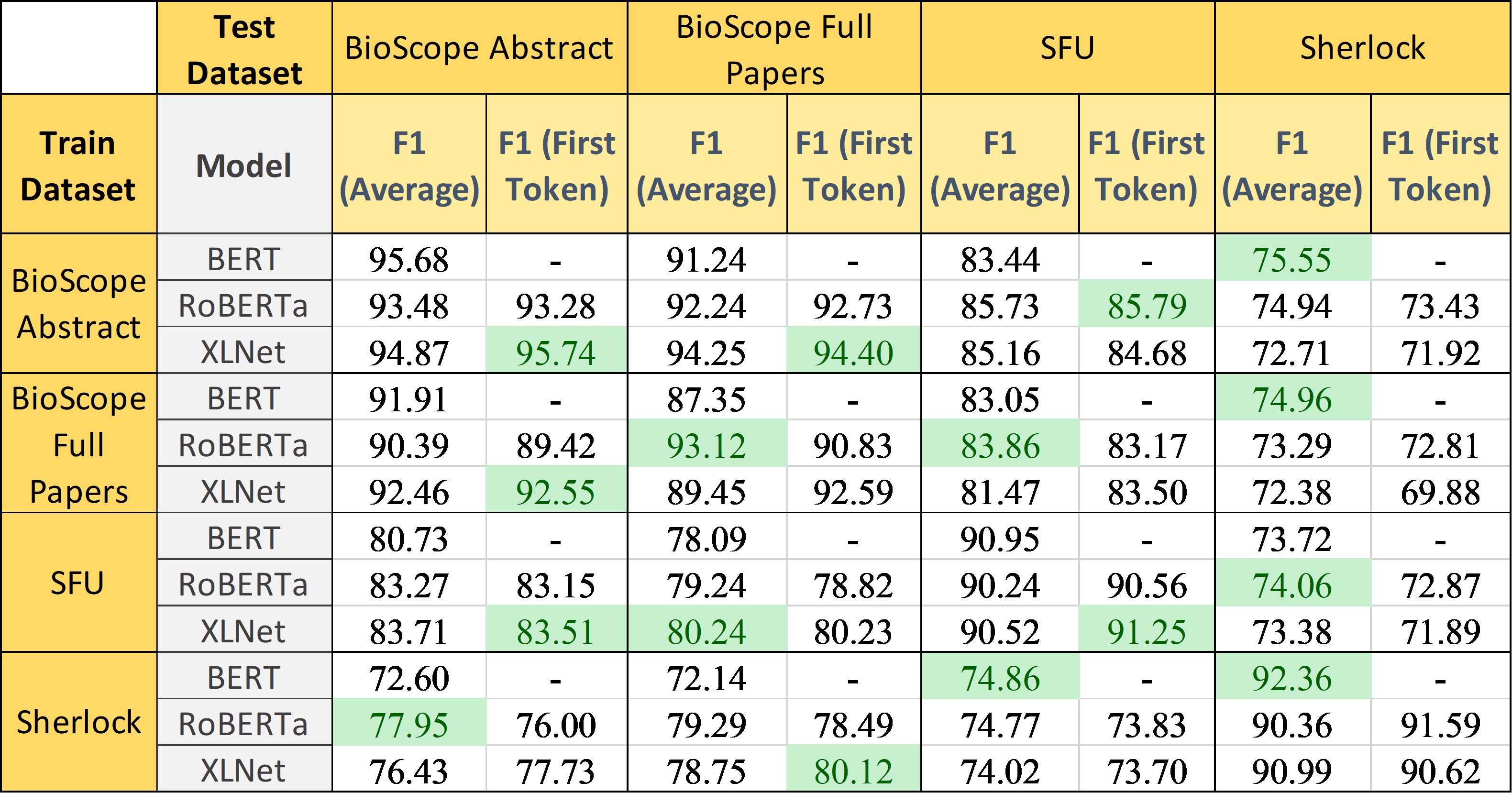}
    \caption{Negation Scope Resolution}
    \label{fig:negscope}
\end{subfigure}

\caption{Results for Negation Detection and Scope Resolution}
\label{fig:neg}
\end{figure}
\noindent
\par We use a default train-validation-test split of 70-15-15 for each dataset, and use all 4 datasets (BF, BA, SFU and Sherlock). The results for BERT are taken from \cite{2019arXiv191104211K}. The results for XLNet and RoBERTa are averaged across 5 runs for statistical significance. Figure \ref{fig:neg} contains results for negation cue detection and scope resolution.
We report state-of-the-art results on negation scope resolution on BF, BA and SFU datasets.
Contrary to popular opinion, we observe that XLNet is better than RoBERTa for the cue detection and scope resolution tasks. A few possible reasons for this trend are:

\begin{itemize}
    \item Domain specificity, as both negation and speculation are closely related subtasks. Further experimentation on different tasks is needed to verify this.
    \item Most benchmark tasks are sentence classification tasks, whereas the subtasks we experiment on are sequence labelling tasks. Given the pre-training objective of XLNet (training on permutations of the input), it may be able to capture long-term dependencies better, essential for sequence labelling tasks.
    \item We work with the base variants of the models, while most results are reported with the large variants of the models.
\end{itemize}

\section{Conclusion and Future Scope}
\noindent
\par In this paper, we expanded on the work of Khandelwal and Sawant (\cite{2019arXiv191104211K}) by looking at alternative transfer-learning models and experimented with training on multiple datasets. 
On the speculation detection task, we obtained a gain of 0.42 F1 points on BF, 1.98 F1 points on BA and 0.29 F1 points on SFU, while on the scope resolution task, we obtained a gain of  8.06 F1 points on BF,  4.27 F1 points on BA and 11.87 F1 points on SFU, when trained on a single dataset. While training on multiple datasets, we observed a  gain of 10.6 F1 points on BF and 1.94 F1 points on BA on the speculation detection task and 2.16 F1 points on BF and 0.25 F1 points on SFU on the scope resolution task over the single dataset training approach. We thus significantly advance the state-of-the-art for speculation detection and scope resolution.
On the negation scope resolution task, we applied the XLNet and RoBERTa and obtained a gain of 3.16 F1 points on BF, 0.06 F1 points on BA and 0.3 F1 points on SFU. 
Thus, we demonstrated the usefulness of transformer-based architectures in the field of negation and speculation detection and scope resolution.
We believe that a larger and more general dataset  would go a long way in bolstering future research and would help create better systems that are not domain-specific.

\bibliographystyle{splncs04}
\bibliography{paper}

\begin{thebibliography}{10}
\providecommand{\url}[1]{\texttt{#1}}
\providecommand{\urlprefix}{URL }
\providecommand{\doi}[1]{https://doi.org/#1}

\bibitem{apostolova-etal-2011-automatic}
Apostolova, E., Tomuro, N., Demner-Fushman, D.: Automatic extraction of
  lexico-syntactic patterns for detection of negation and speculation scopes.
  In: Proceedings of the 49th Annual Meeting of the Association for
  Computational Linguistics: Human Language Technologies. pp. 283--287.
  Association for Computational Linguistics, Portland, Oregon, USA (Jun 2011),
  \url{https://www.aclweb.org/anthology/P11-2049}

\bibitem{diaz-noa-taboada}
Cruz~Diaz, N., Taboada, M., Mitkov, R.: A machine learning approach to negation
  and speculation detection for sentiment analysis. Journal of the American
  Society for Information Science and Technology (JASIST)  (06 2015).
  \doi{10.1002/asi.23533}

\bibitem{DBLP:journals/corr/abs-1810-04805}
Devlin, J., Chang, M., Lee, K., Toutanova, K.: {BERT:} pre-training of deep
  bidirectional transformers for language understanding. CoRR
  \textbf{abs/1810.04805} (2018), \url{http://arxiv.org/abs/1810.04805}

\bibitem{farkas-etal-2010-conll}
Farkas, R., Vincze, V., M{\'o}ra, G., Csirik, J., Szarvas, G.: The
  {C}o{NLL}-2010 shared task: Learning to detect hedges and their scope in
  natural language text. In: Proceedings of the Fourteenth Conference on
  Computational Natural Language Learning {--} Shared Task. pp. 1--12.
  Association for Computational Linguistics, Uppsala, Sweden (Jul 2010),
  \url{https://www.aclweb.org/anthology/W10-3001}

\bibitem{FEI202022}
Fei, H., Ren, Y., Ji, D.: Negation and speculation scope detection using
  recursive neural conditional random fields. Neurocomputing  \textbf{374},  22
  -- 29 (2020). \doi{https://doi.org/10.1016/j.neucom.2019.09.058},
  \url{http://www.sciencedirect.com/science/article/pii/S0925231219313268}

\bibitem{2019arXiv191104211K}
{Khandelwal}, A., {Sawant}, S.: {NegBERT: A Transfer Learning Approach for
  Negation Detection and Scope Resolution}. arXiv e-prints arXiv:1911.04211v3
  (Nov 2019)

\bibitem{kilicoglu-bergler-2010-high}
Kilicoglu, H., Bergler, S.: A high-precision approach to detecting hedges and
  their scopes. In: Proceedings of the Fourteenth Conference on Computational
  Natural Language Learning {--} Shared Task. pp. 70--77. Association for
  Computational Linguistics, Uppsala, Sweden (Jul 2010),
  \url{https://www.aclweb.org/anthology/W10-3010}

\bibitem{konstantinova-etal-2012-review}
Konstantinova, N., de~Sousa, S.C., Cruz, N.P., Ma{\~n}a, M.J., Taboada, M.,
  Mitkov, R.: A review corpus annotated for negation, speculation and their
  scope. In: Proceedings of the Eighth International Conference on Language
  Resources and Evaluation ({LREC}'12). pp. 3190--3195. European Language
  Resources Association (ELRA), Istanbul, Turkey (May 2012),
  \url{http://www.lrec-conf.org/proceedings/lrec2012/pdf/533\_Paper.pdf}

\bibitem{DBLP:journals/corr/abs-1907-11692}
Liu, Y., Ott, M., Goyal, N., Du, J., Joshi, M., Chen, D., Levy, O., Lewis, M.,
  Zettlemoyer, L., Stoyanov, V.: Roberta: {A} robustly optimized {BERT}
  pretraining approach. CoRR  \textbf{abs/1907.11692} (2019),
  \url{http://arxiv.org/abs/1907.11692}

\bibitem{moncecchi-etal-2012-improving}
Moncecchi, G., Minel, J.L., Wonsever, D.: Improving speculative language
  detection using linguistic knowledge. In: Proceedings of the Workshop on
  Extra-Propositional Aspects of Meaning in Computational Linguistics. pp.
  37--46. Association for Computational Linguistics, Jeju, Republic of Korea
  (Jul 2012), \url{https://www.aclweb.org/anthology/W12-3805}

\bibitem{morante-daelemans-2009-learning}
Morante, R., Daelemans, W.: Learning the scope of hedge cues in biomedical
  texts. In: Proceedings of the {B}io{NLP} 2009 Workshop. pp. 28--36.
  Association for Computational Linguistics, Boulder, Colorado (Jun 2009),
  \url{https://www.aclweb.org/anthology/W09-1304}

\bibitem{morante-etal-2010-memory}
Morante, R., Van~Asch, V., Daelemans, W.: Memory-based resolution of
  in-sentence scopes of hedge cues. In: Proceedings of the Fourteenth
  Conference on Computational Natural Language Learning {--} Shared Task. pp.
  40--47. Association for Computational Linguistics, Uppsala, Sweden (Jul
  2010), \url{https://www.aclweb.org/anthology/W10-3006}

\bibitem{ovrelid-etal-2010-syntactic}
{\O}vrelid, L., Velldal, E., Oepen, S.: Syntactic scope resolution in
  uncertainty analysis. In: Proceedings of the 23rd International Conference on
  Computational Linguistics (Coling 2010). pp. 1379--1387. Coling 2010
  Organizing Committee, Beijing, China (Aug 2010),
  \url{https://www.aclweb.org/anthology/C10-1155}

\bibitem{ozgur-radev-2009-detecting}
{\"O}zg{\"u}r, A., Radev, D.R.: Detecting speculations and their scopes in
  scientific text. In: Proceedings of the 2009 Conference on Empirical Methods
  in Natural Language Processing. pp. 1398--1407. Association for Computational
  Linguistics, Singapore (Aug 2009),
  \url{https://www.aclweb.org/anthology/D09-1145}

\bibitem{qian-etal-2016-speculation}
Qian, Z., Li, P., Zhu, Q., Zhou, G., Luo, Z., Luo, W.: Speculation and negation
  scope detection via convolutional neural networks. In: Proceedings of the
  2016 Conference on Empirical Methods in Natural Language Processing. pp.
  815--825. Association for Computational Linguistics, Austin, Texas (Nov
  2016). \doi{10.18653/v1/D16-1078},
  \url{https://www.aclweb.org/anthology/D16-1078}

\bibitem{read-jonathon-velldal}
Read, J., Velldal, E., Oepen, S., Øvrelid, L.: Resolving speculation and
  negation scope in biomedical articles with a syntactic constituent ranker (12
  2011)

\bibitem{ren-yafeng}
Ren, Y., Fei, H., Peng, Q.: Detecting the scope of negation and speculation in
  biomedical texts by using recursive neural network. pp. 739--742 (12 2018).
  \doi{10.1109/BIBM.2018.8621261}

\bibitem{szarvas-etal-2008-bioscope}
Szarvas, G., Vincze, V., Farkas, R., Csirik, J.: The {B}io{S}cope corpus:
  annotation for negation, uncertainty and their scope in biomedical texts. In:
  Proceedings of the Workshop on Current Trends in Biomedical Natural Language
  Processing. pp. 38--45. Association for Computational Linguistics, Columbus,
  Ohio (Jun 2008), \url{https://www.aclweb.org/anthology/W08-0606}

\bibitem{tang-etal-2010-cascade}
Tang, B., Wang, X., Wang, X., Yuan, B., Fan, S.: A cascade method for detecting
  hedges and their scope in natural language text. In: Proceedings of the
  Fourteenth Conference on Computational Natural Language Learning {--} Shared
  Task. pp. 13--17. Association for Computational Linguistics, Uppsala, Sweden
  (Jul 2010), \url{https://www.aclweb.org/anthology/W10-3002}

\bibitem{velldal}
Velldal, E.: Predicting speculation: A simple disambiguation approach to hedge
  detection in biomedical literature. Journal of biomedical semantics
  \textbf{2 Suppl 5}, ~S7 (10 2011). \doi{10.1186/2041-1480-2-S5-S7}

\bibitem{velldal-etal-2010-resolving}
Velldal, E., {\O}vrelid, L., Oepen, S.: Resolving speculation: {M}ax{E}nt cue
  classification and dependency-based scope rules. In: Proceedings of the
  Fourteenth Conference on Computational Natural Language Learning {--} Shared
  Task. pp. 48--55. Association for Computational Linguistics, Uppsala, Sweden
  (Jul 2010), \url{https://www.aclweb.org/anthology/W10-3007}

\bibitem{velldal-etal-2012-speculation}
Velldal, E., {\O}vrelid, L., Read, J., Oepen, S.: Speculation and negation:
  Rules, rankers, and the role of syntax. Computational Linguistics
  \textbf{38}(2),  369--410 (2012). \doi{10.1162/COLI\_a\_00126},
  \url{https://www.aclweb.org/anthology/J12-2005}

\bibitem{Wolf2019HuggingFacesTS}
Wolf, T., Debut, L., Sanh, V., Chaumond, J., Delangue, C., Moi, A., Cistac, P.,
  Rault, T., Louf, R., Funtowicz, M., Brew, J.: Huggingface's transformers:
  State-of-the-art natural language processing. ArXiv  \textbf{abs/1910.03771}
  (2019)

\bibitem{DBLP:journals/corr/abs-1906-08237}
Yang, Z., Dai, Z., Yang, Y., Carbonell, J.G., Salakhutdinov, R., Le, Q.V.:
  Xlnet: Generalized autoregressive pretraining for language understanding.
  CoRR  \textbf{abs/1906.08237} (2019), \url{http://arxiv.org/abs/1906.08237}

\end{thebibliography}
\end{document}